%% file: contextual_motifs.tex
\documentclass[sigconf]{acmart}
\pdfoutput=1 
\usepackage{booktabs} % For formal tables

\usepackage[ruled]{algorithm2e} % For algorithms

\SetAlFnt{\small}
\SetAlCapFnt{\small}
\SetAlCapNameFnt{\small}
\SetAlCapHSkip{0pt}
\IncMargin{-\parindent}
\usepackage[noend]{algpseudocode}
\usepackage{multirow}
\usepackage{subcaption}
\usepackage{graphicx}
\usepackage{enumitem}
\usepackage{framed}
\usepackage{flushend}
\copyrightyear{2017} 
\acmYear{2017} 
\setcopyright{acmlicensed}
\acmConference{KDD '17}{August 13-17, 2017}{Halifax, NS, Canada}\acmPrice{15.00}\acmDOI{10.1145/3097983.3098068}
\acmISBN{978-1-4503-4887-4/17/08} 

\begin{document}
\title{Contextual Motifs}
\subtitle{Increasing the Utility of Motifs using Contextual Data}

\author[*]{Ian Fox}%\authornote{CSE, University of Michigan}}
\affiliation{
  \institution{Computer Science and Engineering, University of Michigan}
  %\streetaddress{P.O. Box 1212}
}
\email{ifox@umich.edu}

\author{Lynn Ang}%\authornote{Department of Internal Medicine, Division of Metabolism, Endocrinology and Diabetes, University of Michigan}}
\affiliation{%
	\institution{Internal Medicine,\\ University of Michigan}
	%\streetaddress{P.O. Box 1212}
}
\email{angly@med.umich.edu}

\author{Mamta Jaiswal}
\affiliation{%
	\institution{Internal Medicine,\\ University of Michigan}
	%\streetaddress{P.O. Box 1212}
}
\email{jaiswalm@med.umich.edu}

\author{Rodica Pop-Busui}
\affiliation{%
	\institution{Internal Medicine,\\ University of Michigan}
	%\streetaddress{P.O. Box 1212}
}
\email{rpbusui@med.umich.edu}

\author{Jenna Wiens}
\affiliation{%
  \institution{Computer Science and Engineering, University of Michigan}
  %\streetaddress{P.O. Box 1212}
}
\email{wiensj@umich.edu}

\renewcommand{\shortauthors}{Fox et al.}

\begin{abstract}
	Motifs are a powerful tool for analyzing physiological waveform data. Standard motif methods, however, ignore important contextual information (\textit{e.g.}, what the patient was doing at the time the data were collected). We hypothesize that these additional contextual data could increase the utility of motifs. Thus, we propose an extension to motifs, \textit{contextual motifs}, that incorporates context. Recognizing that, oftentimes, context may be unobserved or unavailable, we focus on methods to jointly infer motifs and context. Applied to both simulated and real physiological data, our proposed approach improves upon existing motif methods in terms of the discriminative utility of the discovered motifs. In particular, we discovered contextual motifs in continuous glucose monitor (CGM) data collected from patients with type 1 diabetes. Compared to their contextless counterparts, these contextual motifs led to better predictions of hypo- and hyperglycemic events. Our results suggest that even when inferred, context is useful in both a long- and short-term prediction horizon when processing and interpreting physiological waveform data.
	\end{abstract}

%TODO
%\begin{CCSXML}
%<ccs2012>
%  <concept_id>10010520.10010553.10010562</concept_id>
%  <concept_desc>Computer systems organization~Embedded systems</concept_desc>
%  <concept_significance>500</concept_significance>
%  <concept_id>10010520.10010575.10010755</concept_id>
%  <concept_desc>Computer systems organization~Redundancy</concept_desc>
%  <concept_significance>300</concept_significance>
%  <concept_id>10010520.10010553.10010554</concept_id>
%  <concept_desc>Computer systems organization~Robotics</concept_desc>
%  <concept_significance>100</concept_significance>
%  <concept_id>10003033.10003083.10003095</concept_id>
%  <concept_desc>Networks~Network reliability</concept_desc>
%  <concept_significance>100</concept_significance>
%</ccs2012>  
%\end{CCSXML}

%\ccsdesc[500]{Computer systems organization~Embedded systems}

%\keywords{ACM proceedings, \LaTeX, text tagging}

\maketitle

\input{contextual_motifs_body}

\bibliographystyle{ACM-Reference-Format}
\bibliography{sigproc} 

\end{document}

%% file: contextual_motifs_body.tex
\section{Introduction}
In a single day, a continuous glucose monitor can passively collect over 200 times the amount of data one would normally collect using standard blood glucose measurement practices (\textit{e.g.}, manual measurement). Such wearable health monitors have led to the rapid accumulation of large amounts of rich longitudinal health data in the form of time series \cite{nakamura2016multiscale}. Still, in terms of utility, these data have yet to reach their full potential. This is in part do to the fact that we often have an incomplete understanding of the physiological system(s) generating these data. Reasoning about these systems is a central goal of the analysis of physiological time-series data. An increased understanding of these systems could lead to better data-driven predictions and, in turn, better health outcomes. To this end, researchers have sought efficient methods to analyze and represent these time-series data. 

\textit{\textbf{Motifs}}, subsequences that occur in or across signals more often than average, are a popular and useful abstraction for representing and analyzing sequential data \cite{syed2010motif, vahdatpour2009toward}. Motifs are short subsequences or patterns that commonly occur within or across signals. For example, we might \textit{discover} a motif representing a rapid heartbeat within an electrocardiogram (ECG) dataset. By representing an ECG signal by whether or not it contains this particular motif, we could learn an association between a rapid heartbeat and a patient's cardiac health (\textit{e.g.}; a rapid heartbeat may be indicative of poor health).

Motifs exhibit several properties that make them an appealing approach for analyzing large physiological datasets, including their broad applicability across a wide range of data types, their resilience to noise, and their inherent interpretability \cite{bailey_fitting_1994, breton2008analysis, chui2014introduction, mueen2014time, syed2010motif}.

Despite their useful properties, Van Esbroeck et al. recently showed that, as a feature representation for discriminative algorithms, motifs performed worse than other representation learning schemes \cite{van2015learning}. We hypothesize that a key reason for this suboptimal performance is that standard motifs ignore the \textit{context} under which they are generated. Standard motif methods assume that the system which generates the signal, and thus the meaning of the motifs, does not change over the observation period. This assumption is not always reasonable in physiological data.

For example, suppose that the ECG data mentioned earlier were collected under a range of normal living conditions. The ability of the heart rate to vary to accommodate external stressors, called heart rate variability, indicates good cardiac health \cite{kleiger1987decreased}. Rapid heartbeats would then be an indicator of good health within the context of physical activity, and an indicator of poor health outside that context. 

Another, particularly relevant, example can be found in glucose data (\textit{i.e.}, blood sugar levels). In this domain, motifs may represent spikes in blood glucose levels. When interpreting these spikes, it is important to understand the broader context of the signal. Occasional, isolated spikes may be expected as the result of glycemic challenges (such as meals). In contrast, repeated spikes in a temporally localized region could indicate poor glycemic control, a risk factor for long-term health outcomes \cite{jaiswal2014association, muggeo2000fasting}. Thus, it is not always reasonable to assume that the interpretation of a motif remains constant throughout the signal, especially in physiological data collected under varying conditions. 

We address these issues by augmenting motifs with context. Here, we define context as external phenomena affecting the signal, \textit{e.g.}, physical activity in ECG, or meals in glucose data. We present a framework, for understanding physiological time series, that incorporates time-varying context. By adding context, we hypothesize that we can more accurately model the relationship between the signal and outcomes of interest. Importantly, context can be either observed or unobserved. In our blood glucose example, if patients' meals were never recorded, the context would be unobserved. In this common setting, accounting for context is more challenging. In light of these challenges, we:
\begin{enumerate}
	\item propose a general motif framework that accounts for context, \textit{\textbf{contextual motifs}}, and
	\item introduce contextual motif discovery along with techniques to jointly infer context and motifs in a signal when context is unobserved.
\end{enumerate}

One could discover contextual motifs using a two-stage approach in which context and motifs are inferred independently. However, we hypothesize that a joint inference approach will lead to richer motifs, as it allows contextual information to improve motif discovery while simultaneously using motifs to improve context discovery.

Applied to both simulated and real physiological data, our proposed approach improves upon existing motif methods in terms of the discriminative utility of the discovered motifs. We observe a consistent 11-12 percentage point improvement in the area under the ROC curve (AUC) using contextual motifs compared to a contextless baseline, across a range of settings in simulated data. Additionally, applied to a Continuous Glucose Monitoring (CGM) dataset of patients with type 1 diabetes (T1D), our proposed contextual motif discovery techniques led to a 4 and 7.2 point improvement in the AUC when predicting long-term hypo- and hyperglycemic events respectively, compared with a contextless motif discovery method. We demonstrate that our method is capturing a physiologically relevant context for glucose data, demonstrating the potential of joint motif-context inference. These results suggest that, even when context is unobserved, contextual motifs could lead to a better understanding of the physiological system and, therefore, more accurate predictions.  
 
\section{Background and Problem Statement}
Motifs have been applied across many different fields, including: genetic analysis, activity monitoring, and clinical event prediction \cite{liu2015efficient, minnen2006discovering, zhou2004cismodule}. For an in-depth review of motifs and their applications, we refer the reader to the survey paper by Mueen \cite{mueen2014time}. Here, we limit our discussion to prior work that directly relates to our proposed approach. We introduce vocabulary for discussing different motif discovery approaches. In particular, we differentiate between \textit{data-derived} and \textit{data-generating} motifs. Then, we present our proposed framework, \textit{contextual motifs}, and discuss related work.

\subsection{Motifs}
\subsubsection{History, Definitions, and Notation}
The study of motifs originated in the field of genetics \cite{needleman1970general}. Lin et al. ported the idea of motifs to time-series analysis \cite{lin2002finding}. While there exist several different definitions of time-series motifs, here, we focus on \textit{support motifs} \cite{mueen2014time, syed2010motif}. Support motifs are the subsequences that are most frequently expressed throughout a dataset \cite{mueen2014time}. These types of motifs are best suited to capturing inter- and intra-sequence similarity, and are widely used as a result \cite{buhler2002finding, grabocka2016latent, jensen2006generic, liu2015efficient, minnen2006discovering, oates2002peruse}.

Before going further, we provide some notation and definitions to make our discussion more precise. We define a \textit{signal} as an ordered observation sequence of the form $\mathbf{x} = (x_1, x_2, \dots, x_T)$, where $T$ is the signal length. Signals can be any form of ordered data, but here we focus on data ordered by time, also called \textit{time series}. A \textit{subsequence} of a signal is a contiguous subset of observations $\mathbf{x}_{t,k} = [x_t, x_{t+1}, \dots, x_{t+k-1}]$, here the subsequence has length $k$. We call the process that generates the signal the \textit{system}. Our goal in time-series analysis is to increase our understanding of the system using the observed signal. Some subsequences help us achieve this goal, whereas others do not. Subsequences that occur more often (\textit{i.e.}, have support) are less likely to be the result of noise, and thus are more likely to reflect structure in the system. This motivates the study of frequently occurring subsequences, or \textit{motifs}.

Previous work has investigated how to increase the discriminative power of motifs. Motifs that are maximally discriminative across classes are known as shapelets \cite{ye2009time}. Shapelets extend motifs by using a measure of their predictive power during the discovery step. Other work has sought to discover more discriminative patterns by directly using label information \cite{cheng_discriminative_2007, ranu_graphsig:_2009}. Our work also extends motif discovery to better predict outcomes. However, we approach this problem in a unsupervised manner. I.e., we focus on learning a richer representation of the signal, leading to the discovery of more informative motifs without using labels. Our approach is complementary to previously proposed supervised methods. 

\subsubsection{Data-Derived vs. Data-Generating Motifs} Data-derived and data-generating motifs are the two most common frameworks in the motif literature. Data-derived motifs are subsequences that are \textit{derived} from the signal. Data-generating motifs are latent variables that \textit{generate} the signal. See Table \ref{tab:dd_dg} for a summary of data-derived versus data-generating motifs.

\textbf{Data-derived motifs} are discovered using a measure of subsequence quality $q$. A motif $\mathbf{m}_t$ is a subsequence 
$$\mathbf{m}_t = \mathbf{x}_{t,k}: q(\mathbf{x}_{t,k}) > n$$ 
for some quality threshold $n$. With support motifs, the quality measure $q$ indicates the number of \textit{approximate} occurrences of the subsequence in a dataset. Looking for exactly repeated subsequences is limiting, because noise can perturb even well conserved subsequences. Thus, the vast majority of motif discovery procedures attempt to find approximate motifs.

\textbf{Data-generating motifs} are discovered using a generative model to encode assumptions about how the data were created. These motifs are defined as an ordered sequence of distributions $$\mathbf{m}_z = [\boldsymbol{\theta}_{z, 1}, \boldsymbol{\theta}_{z, 2}, \dots, \boldsymbol{\theta}_{z, k}] : \boldsymbol{\theta}_{z, i} \in \boldsymbol{\Theta} \text{ for }i=1, \dots, k$$ where elements of $\boldsymbol{\Theta}$ parameterize the distribution generating the signal and $z$ is the motif label. Note that under this definition, a motif is not parameterized by the values occurring in the signal, but instead by a sequence of distributions. 

The primary difference between these motif types is the method by which they are discovered. Data-derived motif discovery methods focus on search. In contrast, data-generating motif discovery methods emphasize inference. Both approaches are useful to consider. Though data-derived motifs are easier to discover and thus more common in the literature \cite{buhler2002finding, lin2002finding, liu2015efficient, mueen2014time, shokoohi2015discovery}, data-generating motifs (also called latent motifs), are typically higher quality \cite{grabocka2016latent, saria_discovering_2011}. One can naturally extend discovering data-generating motifs to discovering deformable motifs, motifs that have similar structure but vary in duration. This can be particularly relevant in the analysis of physiological data \cite{saria_discovering_2011}.

\begin{table}
	\caption{Summary of Data-Derived vs. Data-Generating Motifs. Both are important types of motifs, and we consider both in developing contextual motifs.}
	\label{tab:dd_dg}
	\begin{tabular}{clll}
		\toprule
		 & Discovery & Advantages & Baseline\\
		\midrule
		Data-Derived & search & efficiency & MDLats \cite{liu2015efficient} \\
		Data-Generating & inference & quality & MMM \cite{bailey_fitting_1994} \\
		\bottomrule
	\end{tabular}
\end{table}

\subsubsection{Discovering Motifs - Baseline Methods}
\label{sec:base}
We consider two motif discovery algorithms as baselines. The first is a data-derived motif discovery algorithm, MDLats. Presented by Liu et al. \cite{liu2015efficient}, MDLats combines many of the methodological improvements developed in recent literature into an efficient and effective method of discovering support motifs in physiological signals \cite{buhler2002finding, chiu2003probabilistic, liu2015efficient}. We will show how MDLats can be applied in a contextual motif framework to further improve discriminative performance.

The second baseline we consider, based on a data-generating motif discovery algorithm proposed by Bailey et al., is a motif mixture model (MMM) \cite{bailey_fitting_1994}. Their method is designed for categorical genomic data, and thus fits a mixture of categorical distributions. As we are working with real-valued physiological time series, we instead use a Gaussian mixture model. We assume each portion of a signal is generated by a set of distributions called a motif. For one signal, $\mathbf{x}$, the generative model is given below: 
\begin{enumerate}
	\item For $i=0, \dots, K$:
	\begin{enumerate}
		\item Draw motif label $z_i \sim  \text{Cat}(\boldsymbol{\gamma})$, where $\boldsymbol{\gamma}$ is the motif mixing parameter
		\item For $k=0, \dots, l_m-1$: 
		\begin{enumerate}
			\item Draw observation: $$x_{i(l_m)+k} \sim \mathcal{N}(\boldsymbol{\theta}_{z_i, \mu_k}, \boldsymbol{\theta}_{z_i, \sigma^2_k})$$ Where $K$ is the signal length divided by the motif length, and $\boldsymbol{\theta}_{z_i}$ parameterizes the $K$ component distributions for the motif $\mathbf{m}_{z_i}$
		\end{enumerate}
	\end{enumerate}
\end{enumerate}

Thus to generate a signal, one draws a categorical motif label $z$ from a categorical distribution parameterized by $\boldsymbol{\gamma}$. This motif operates over some period of time called the motif window with length $l_m$. Observations from the signal are then distributed according to the motif-dependent parameters $\boldsymbol{\theta}_{z}$.

According to this definition, even noisy, poorly conserved subsequences are technically motifs. Thus, it is common to include a \textit{background} motif, assumed to be poorly conserved, to account for such subsequences. Here, inferring $\boldsymbol{\gamma}$ and $\boldsymbol{\theta}_{z}$ for each motif $\mathbf{m}_z$ is identical to inference with a Gaussian mixture model. Each mixture component, representing a motif, is an $l_m$-dimensional Gaussian with a diagonal covariance matrix.

\subsection{Contextual Motifs- A Novel Extension}
\label{subsec:CM}
We extend past work on motifs to incorporate context. We represent context as a categorical variable $c_t \in C$ that, at time point $t$, takes a discrete value $c_t \in \{1, \dots, n_c\}$, where $n_c$ is the number of distinct contexts. Contextual motifs are then well conserved patterns of contextualized data, representing a motif and the context under which it occurs. Contextual motif discovery is the task of discovering these contextual motifs. Discovery can also be viewed as the process of discovering motifs occurring within similar contexts. This distinction can be important when contexts vary from signal to signal. Without taking context into account, our measure of motif quality $q$ may mistake infrequent contexts with infrequent motifs within a context. For example, a certain quantity of food consumption may always present a distinctive blood glucose pattern in individuals with poor glycemic control. If some glucose signals contained large meals, and others did not, contextless motif discovery could fail to recognize this pattern. When context is observed, as a context signal $\mathbf{x}_C$, discovering contextual motifs is trivial. One can simply apply an existing motif discovery algorithm to each chunk of data occurring under a particular context. Contextual motifs in this simple setting would be tuples $(\textbf{m}_t, c_i)$.  

A more interesting, and perhaps more common, setting is the one in which context is unobserved or unavailable. In the next section, we focus on methods for discovering contextual motifs in this latent setting. We present methods that extend both data-derived and data-generating motifs to contextual motifs. When extending data-derived motifs, we infer context in a two-stage approach. While such an approach could also apply to data-generating motifs, we sought a joint inference approach, in which context and motifs inform one another. Data-generating motifs can be extended using either a two-stage approach or by jointly inferring motifs and context. To perform this joint inference, we propose a generative model based on a subclass of dynamic Bayesian networks \cite{murphy2002dynamic}.

Despite the vast literature studying motifs, and motif discovery methods, there is relatively limited work on considering abstractions based on motifs. Still, we discuss the related work in this area and how it differs from what we propose.

Van Esbroeck et al. considered representing physiological signals using a bag of motifs (a common approach in motif discovery) \cite{van2012heart}. But, the authors built upon this representation using a topic modeling approach. In their approach, each topic is associated with a distribution over motifs and each signal is then a distribution over topics. These topics are then used as an abstraction to represent the signal. While the idea of a topic is conceptually similar to context, their method differs from our proposed method in two ways. First, we jointly infer contexts and motifs, whereas they assume that motifs are first discovered and then topics are inferred. Second, the bag-of-motifs approach assumes a static representation of the signal. In contrast, our approach leverages the temporal contiguity of contexts to allow for flexible variation in motif representations both within and across signals.

In other domains, where the order in which motifs occur is important (\textit{e.g.}, genetics) researchers have considered leveraging this temporal ordering to discover better motif representations. In particular, Lin et al. proposed a method based on hierarchical HMMs in which the presence/absence of motifs in one part of a signal affects the presence/absence of motifs in neighboring parts \cite{lin2008baycis}. Again, this is conceptually similar to context, since context can be viewed as a type of inter-motif structure. However, like the method described above, Lin et al.'s method also requires a separate motif discovery step. Our approach combines the discovery steps, enabling joint motif-context discovery.

Finally, others have looked at non-motif based approaches to learning abstractions based on time-series data. For example, Saria et al. examined the joint discovery of generating functions and temporal topics in NICU data \cite{saria2010learning}. They also used a hierarchical model to represent structure at different temporal levels. However, they assumed the use of base autoregressive generating functions. We instead use motifs as our primitive of choice, for the reasons described above. 

\section{Methods}
In this section, we propose approaches for discovering contextual motifs when context is unobserved (\textit{i.e.}, latent). In this setting, we begin with a straightforward extension of data-derived motifs to contextual motifs using a two-stage approach. Then, we move on to the more interesting setting, in which we show how to extend data-generating motifs to contextual motifs. In particular, we present a method to jointly discover motifs and context. 

\subsection{Two-Stage Discovery with Data-Derived Contextual Motifs}
\label{sec:two_stage}
By adopting a two-stage approach, we can adapt any existing motif discovery method (data-generating or data-derived) to discover contextual motifs. We can use any context discovery algorithm appropriate to our problem to derive the context signal $\mathbf{x}_C$, followed by the application of a motif discovery algorithm within each context. We note that the motif and context discovery steps in the two-stage approach can be performed in either order.

\begin{figure}
	\centering
	\includegraphics[width=1\linewidth]{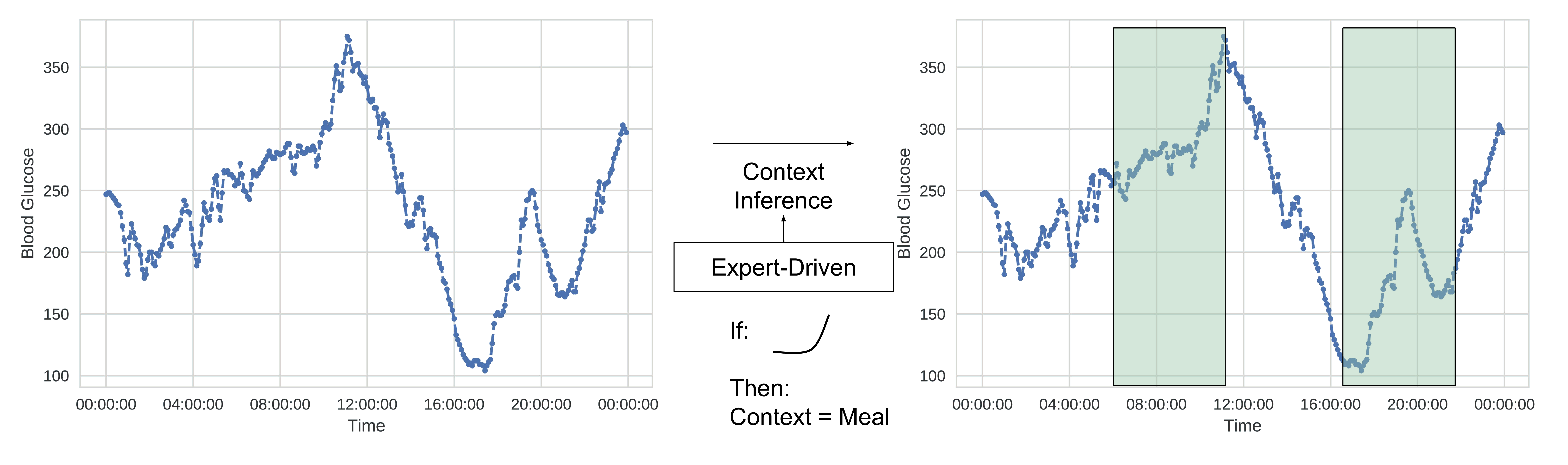}
	\caption{When expert/domain knowledge is available, we can discover context using hand-engineered rules. The rule illustrated here assigns a `meal' context to a window (highlighted in green) around a sharp increase in blood sugar.}
	\label{fig:ContextInference}
\end{figure}

In our domain of interest, predicting outcomes from blood glucose, two plausible context discovery approaches include:
\begin{enumerate}
	\item Expert-Driven Context: Using domain knowledge we can hand-engineer features indicative of certain contexts. For example, large spikes in blood glucose levels typically occur only after meals \cite{MAGE, molnar1970mean}. Using this knowledge, we created an expert rule,
	\begin{gather*}
	c_t = b(\frac{\sum_{i=1}^{k} x_{t-(i-1)} - x_{t-i}}{k})
	\end{gather*}
	where $b$ is a hand-defined mapping from $\mathbb{R} \rightarrow C$. Figure \ref{fig:ContextInference} illustrates the process of context discovery.
	
	\item Data-Driven Context: Using an unsupervised method, an HMM, we infer transient structure in the data. The hidden states then are used as context.
\end{enumerate}

Alternatively, motifs can be used to derive context. If we observe the motif labels $\mathbf{m}$, found via an initial motif discovery step, then we can cluster empirical frequencies within the context windows to derive the contextual motif distribution parameter $\boldsymbol{\gamma}$ and use maximum likelihood estimation to find $\mathbf{c}$. This is similar to the motif topic context introduced by \cite{van2012heart}, except under this approach context varies over the course of the signal.

While straightforward and flexible, this two-stage approach misses an opportunity to learn richer motifs based on context and vice versa. In the activity monitoring literature it has been noted that motifs are useful for deriving context\cite{minnen2006discovering}, and as we will demonstrate with our two-stage discovery experiments, context aids motif discovery. This observation inspired our proposed joint inference approach, presented in the next section.

\subsection{Joint Discovery with Data-Generating Contextual Motifs}
\begin{figure}
	\centering
	\includegraphics[width=1\linewidth]{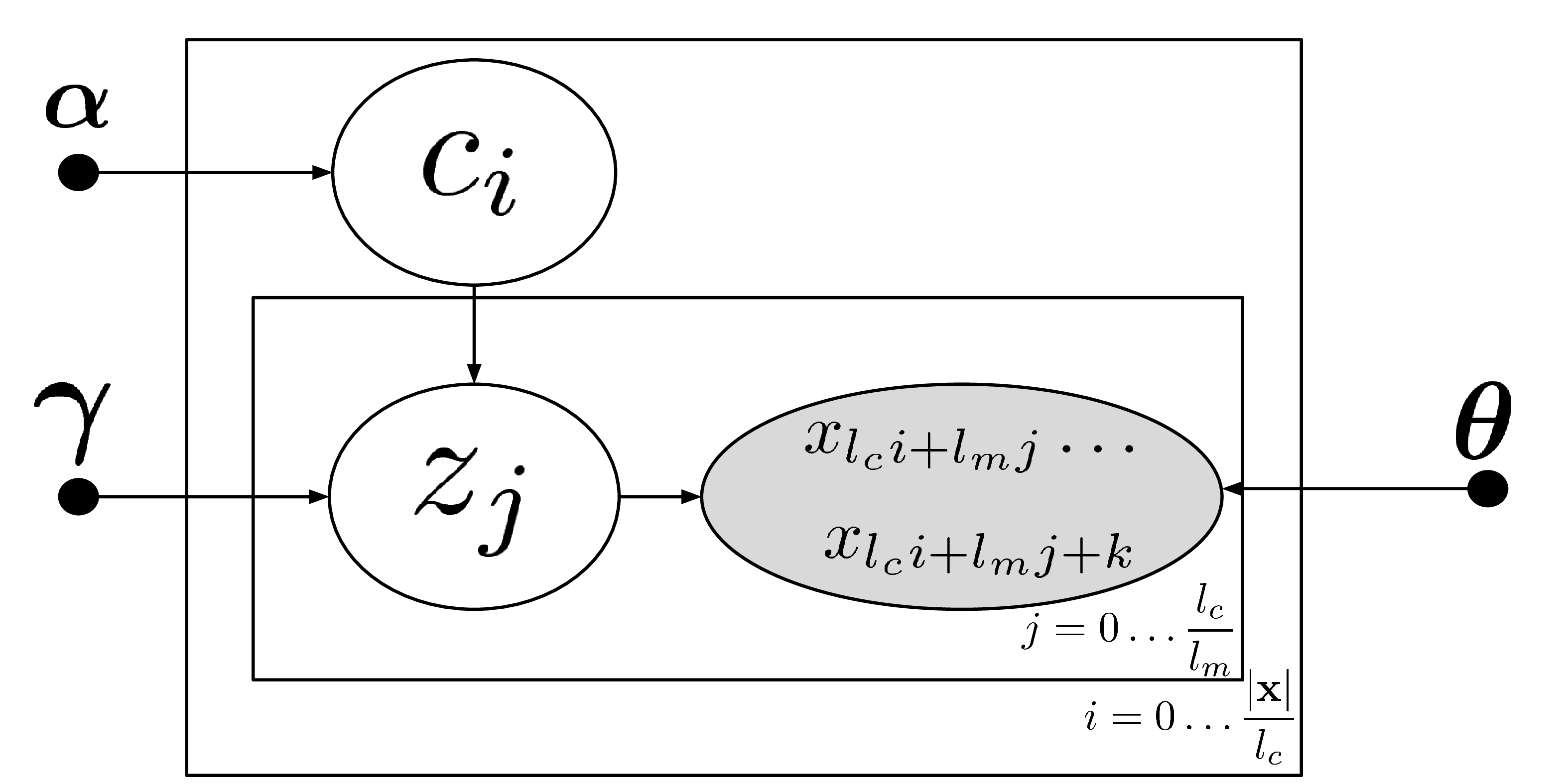}
	\caption{Proposed generative model for the contextual motif generative process. Each context parameterizes a different distribution of motif frequencies.}
	\label{fig:cmmm}
\end{figure}

While data-generating motif discovery techniques carry additional computational cost, the increased model flexibility allows for the joint discovery of motifs and context. To achieve this, we modify the data-generating MMM (introduced in Section \ref{sec:base}) to include a context variable. In our proposed \textit{contextual motif mixture model} (CMMM), each context $c_i$ defines a distribution over the set of possible motifs (Figure \ref{fig:cmmm}). This definition of context is motivated by previous work done in topic models, where the relative frequencies of primitives (originally, words) gives insight into higher level structure in the data \cite{blei2003latent}. The CMMM generative model is as follows:
\begin{enumerate}
	\item For $i=0, \dots, \frac{|\mathbf{x}|}{l_c}$, where $l_c$ is the context window length:
	\begin{enumerate}
		\item Pick context $c_i \sim  \text{Cat}(\boldsymbol{\alpha})$, where $\boldsymbol{\alpha}$ is the context mixing parameter
		\item For $j=0, \dots, \frac{l_c}{l_m}$:
		\begin{enumerate}
			\item Choose motif label $z_j \sim  \text{Cat}(\boldsymbol{\gamma}_{c_i})$, where $\boldsymbol{\gamma}_{c_i}$ is the motif mixing parameter for context $c_i$
			\item For $k=0, \dots, l_m-1$:
			\begin{enumerate}
				\item Draw observation: $$x_{l_c i+l_m j+k} \sim \mathcal{N}(\boldsymbol{\theta}_{z_j, \mu_k}, \boldsymbol{\theta}_{z_j, \sigma_k}^2)$$
			\end{enumerate}
		\end{enumerate}
	\end{enumerate}
\end{enumerate}

Generating data using this model begins by picking a categorical context $c_i$ that holds for some window in time. If one assumes that context changes slowly relative to the rate of data sampling, context windows may be large. Contexts are selected according to a categorical distribution parameterized by a context mixing parameter $\boldsymbol{\alpha}$. 

The selection of a context decides the motif frequency parameter $\boldmath{\gamma}_{c_i}$, and from this point the model behaves identically to the MMM discussed in Section \ref{sec:base}. This model can be viewed as a hierarchical HMM with a feed-forward context layer. Our notion of context is a temporally contiguous extension of the topics used in the motif topic model of Van Esbroeck et al. \cite{van2012heart}. In contrast to their approach, we allow the motif distribution parameter, $\boldsymbol{\gamma}$, to vary over time according to a categorical context variable. We explicitly model motif mixing distributions for each context, but hold the motif parameters constant across all contexts. This allows for the identification of similar contexts across a population.

Using this model, one can infer the specific contexts under which each motif occurs at the same time one infers the motifs, allowing for the \textit{de novo} discovery of contextual motifs.

Discovering the contextual motif representation of a signal requires inferring the motif labels $\mathbf{z} = [z_{j_0}, z_{j_1}, \dots, z_{j_{|\mathbf{x}|/l_m}}]$ and context labels $\mathbf{c} = [c_{i_0}, c_{i_1}, \dots, c_{i_{|\mathbf{x}|/l_c}}]$. This is a mixture of mixture models, and thus exact inference on it is intractable. Note that computing the posterior of the latent variables and parameters requires computing:
\begin{gather*}
\frac{p(\mathbf{c}, \mathbf{z}, \boldsymbol{\alpha}, \boldsymbol{\gamma}, \boldsymbol{\theta}, \mathbf{x})}{\int_{\boldsymbol{\alpha}} p(\boldsymbol{\alpha}) \sum_{\mathbf{c}} p(\mathbf{c}|\boldsymbol{\alpha}) \int_{\boldsymbol{\gamma}} p(\boldsymbol{\gamma}) \sum_{\mathbf{z}} p(\mathbf{z}|\boldsymbol{\gamma}, \mathbf{c}) \int_{\boldsymbol{\theta}} p(\boldsymbol{\theta}) p(\mathbf{x} | \boldsymbol{\theta}, \mathbf{z})}
\end{gather*}

To perform approximate inference on this model, we use a sampling approach, as is routine with data-generating motif systems \cite{lin2008baycis, siddharthan2005phylogibbs, syed2010motif}. Our sampling method was implemented using the probabilistic programming python module PyMC3 \cite{salvatier2016probabilistic}. We sample on the graphical model proposed in Figure \ref{fig:cmmm}. We use different sampling methods on the various types of variables in the model to optimize sampling speed and per-sample improvement. We sample the categorical variables $\textbf{c}$ and $\textbf{z}$ using a Metropolized Gibbs sampler with a uniform proposal distribution \cite{liu1996metropolized}. The motif and context mixing parameters, $\boldsymbol{\alpha}$ and $\boldsymbol{\gamma}$, are assumed to have Dirichlet priors to allow for collapsed sampling. They are sampled using Metropolis-Hastings with a Gaussian proposal distribution. Finally, we sample the motif distribution parameter $\boldsymbol{\theta}$ using a No-U-Turn Sampler with dual averaging \cite{hoffman2014no}. This sampler uses gradient information to intelligently guide the sampling steps, and features an adaptive stepsize. These sampling schemes use the default parameters in PyMC3.

To avoid label switching problems when sampling, we use a series of potentials that enforce an ordering on the dimensions of the variable $\boldsymbol{\alpha}$. We also enforce that values of $\boldsymbol{\alpha}$ are above a threshold dependent on $n_c$, to ensure multiple contexts are used. We draw 2000 samples from the model, where each step iterates over all variables, and derive values using the final 1000. 

\section{Experiments and Results}
\label{sec:exp}
 To measure the utility of contextual motifs, we compare their discriminative performance to that of contextless motifs across several classification tasks.  We also measure the utility of the joint context-motif inference approach compared to a two-stage approach. In the sections that follow, we begin by describing the data used for these experiments and our evaluation procedure. We then present a series of experiments using both data-derived and data-generating motif frameworks on real and simulated physiological data. 

\subsection{Dataset and Prediction Tasks}
\label{sec:cd}
\subsubsection{CGM Data}
Our physiological dataset consists of CGM sessions collected from 40 T1D patients at approximately 3-month intervals over the course of 3 years. The length of each session varies from 3-6 days with glucose measured every 5 minutes. Patients were blinded to their CGM monitor during this time. These monitors, by measuring glucose concentrations in interstitial fluid, accurately infer blood glucose levels between 40-400 mg/dL. The collection and study of these data was approved by a UM Medical School IRB panel.

\begin{figure}
	\centering
	\includegraphics[width=1\linewidth]{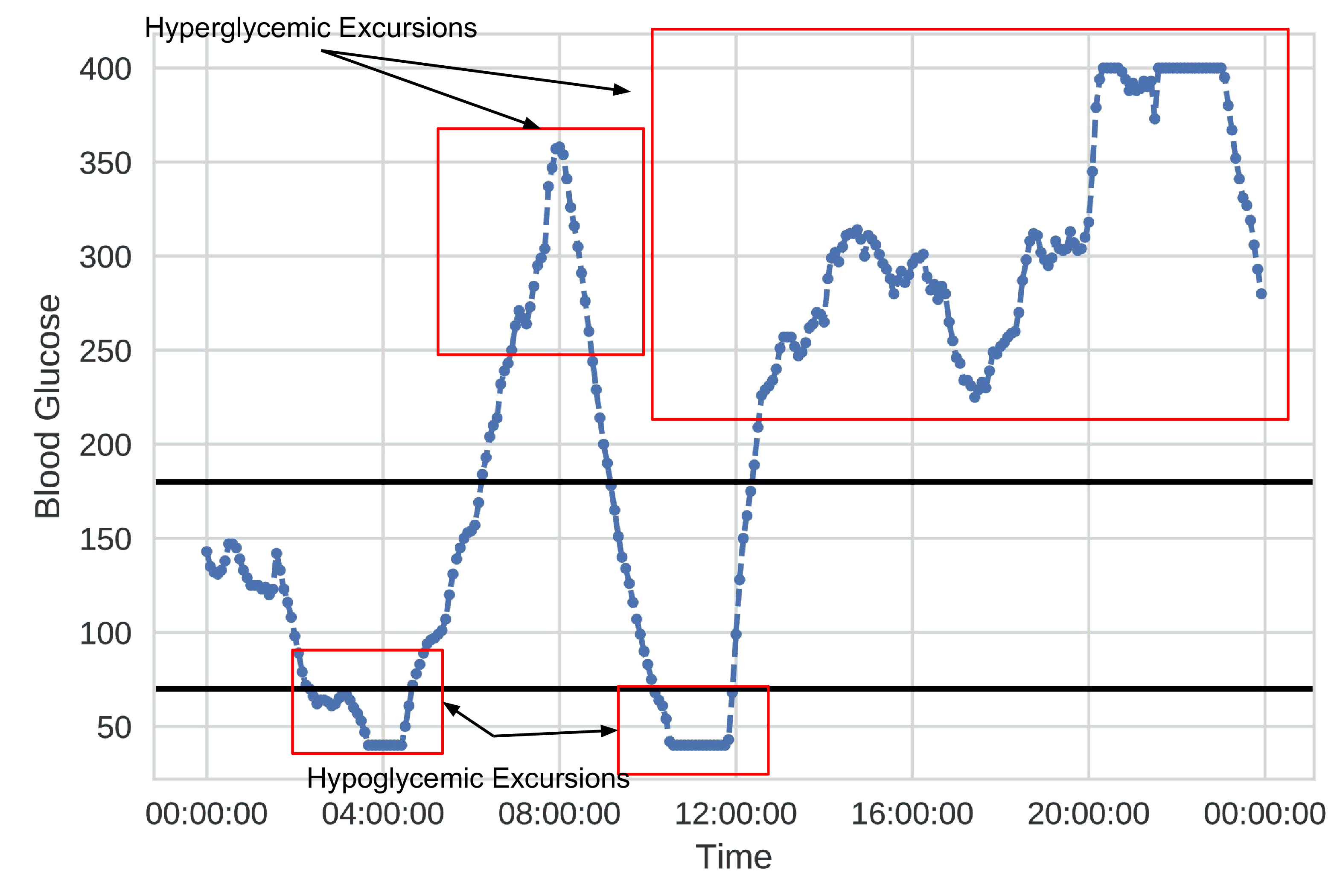}
	\caption{One day of blood glucose in the CGM dataset. Each of the 40 patients has between 21-48 days of data.}
	\label{fig:ExampleofCGMdata}
\end{figure}

Due to calibration and sensor errors, the sessions contain periods of missing data. In total, the dataset contains approximately 51.6k hours of glucose data, and is missing around 4.5k hours. To handle these missing data, we use linear interpolation. This is sensible for small chunks of missing data, as glucose data are locally linear. However, we exclude days with contiguous missing periods longer than thirty minutes. After removing days with excessive missing data, 32.4k hours of data occurring across 1,352 days remain. Each of the 40 patients contributes between 21 and 48 days to this total.

\subsubsection{Evaluation Metric.}
\label{sec:eval}
To measure the utility of the discovered contextual motifs relative to motifs, we represent each day's worth of data by the number of times each (contextual) motif appears in the signal that day, and use this representation in a supervised learning task. \textit{I.e.}, we transform each signal into a feature vector representing motif counts. Given the fixed length feature representation, any number of machine learning techniques can be used to learn a mapping from the feature vector to the label. 

We consider two different supervised learning tasks related to T1D: hypoglycemia (low blood sugar) and hyperglycemia (high blood sugar). Hypoglycemic events are caused by an over administration of insulin, a mismatch in insulin delivery and food absorption, or increased physical activity. They can result in unconsciousness, seizures, irregular heartbeat, and even death \cite{zoungas2010severe}. Hyperglycemic events are caused by overconsumption of carbohydrates or an under administration of insulin. In the short term, severe hyperglycemic events can result in coma or can lead to life threatening diabetic ketoacidosis. In the long term, chronic hyperglycemia increases risk for heart disease, nerve damage, kidney damage, and early death \cite{bergenstal2015glycemic}.

We selected these two tasks for their clinical relevance, but also since they differ substantially in terms of the underlying pathophysiology. Learning a representation that works well across both tasks is more challenging, but ultimately, the goal of this work.

\begin{figure}
	\centering
	\includegraphics[width=1\linewidth]{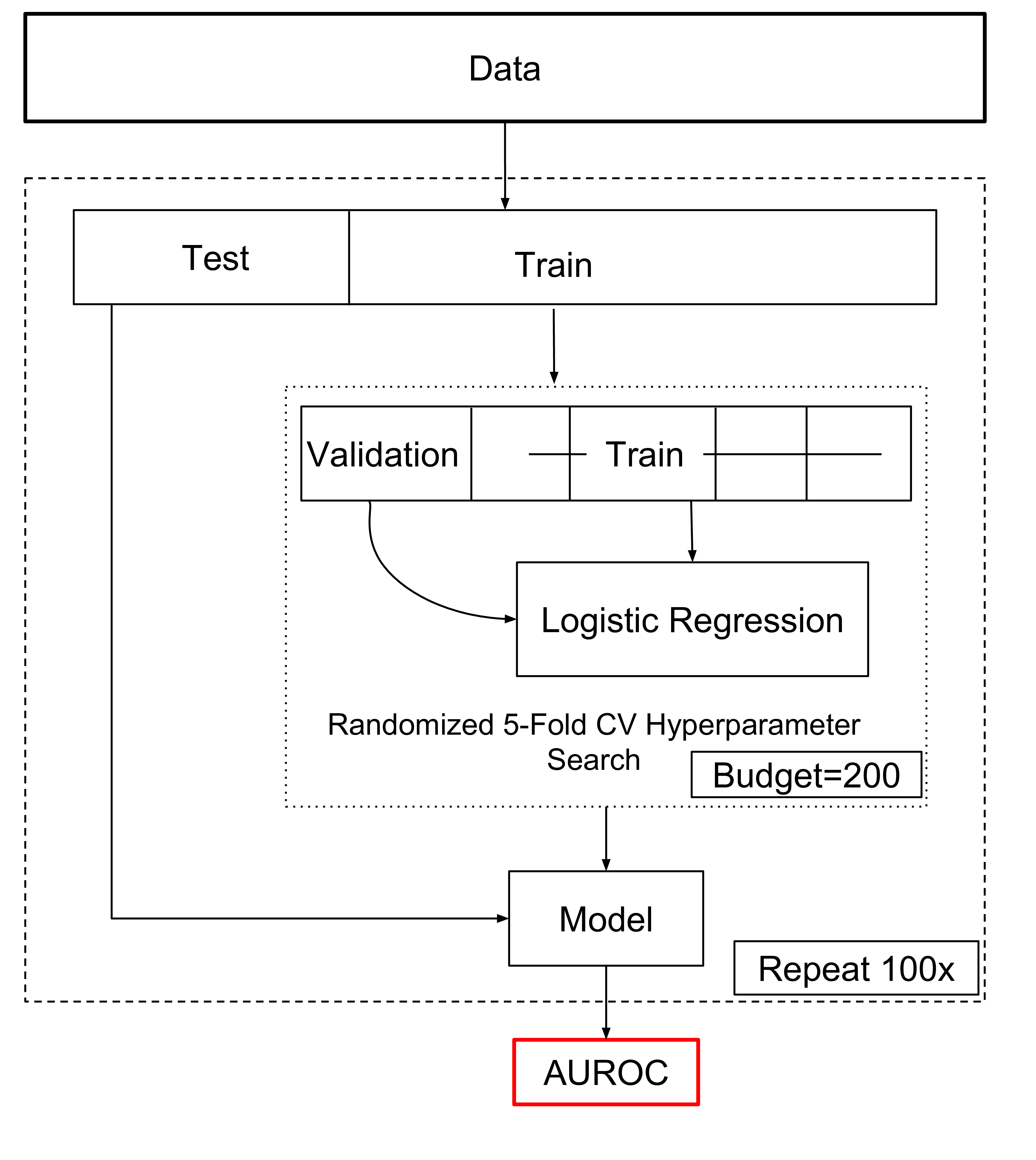}
	\caption{Outline of evaluation process. Data are split, using a patient aware scheme, into train-test splits 100 times. Within each split, the training data are used For model selection. The final classifier is then applied to the test data and the AUC is calculated. We report average AUC across all test sets. }
	\label{fig:cross_val}
\end{figure}

We investigate performing these tasks over both short- and long-term prediction horizons. For the long-term horizon, we use one day of data to predict whether or not a hypo- or hyperglycemic event will occur in the next day. Due to the large number of hyperglycemic events that occur in our data, we modified the prediction task to predict the occurrence of two or more hyperglycemic events. This task measures our ability to represent a patient's long-term physiological state, giving us a measure of their level of glucose control, and is very challenging to do using only one day of data. For the short-term horizon, we use one day of data to predict if a hypo- or hyperglycemic event will occur in the next 40 minutes. This task was chosen to emphasize the immediate state of the patient, and is clinically motivated by the ability of such short-term prediction to aid closed loop glucose control using an artificial pancreas \cite{weinzimer_fully_2008}.

We do not focus on the choice of supervised learning algorithm, as our contribution lies in the feature representation method. For consistency, we use a pipeline constructed using sklearn \cite{pedregosa2011scikit} that tunes hyperparameters and fits a regularized logistic regression classifier using patient-aware stratification for cross-validation and train-test splits. An outline of our evaluation process is given in Figure \ref{fig:cross_val}. We report results in term of average AUC across test sets.

\subsection{Experiments on Real Data}
We begin with the results of our experiments on the CGM dataset. We examine the utility of our data-derived contextual motifs, followed by an evaluation of the proposed data-generating contextual motifs.

\subsubsection{Data-Derived Motifs}
\label{sec:iiexp}

\begin{table}
	\caption{Predictive utility of contextual data-derived motifs. Across both short- and long-term prediction horizons we observe an increase in AUC when including expert- and data-driven contextual information, though the data-driven context is more informative. Bold values indicate the best in prediction class.}
	\label{tab:dd_res}
	\begin{tabular}{ccll}
		\toprule
		\multicolumn{1}{c}{Prediction Task} & \multicolumn{1}{c}{Method} & \multicolumn{2}{c}{AUC (std)}\\
		\midrule
		& Context & Hypo & Hyper\\
		\midrule
		\multirow{3}{*}{Long-Term} &None & 0.535 (0.022) & 0.527 (0.020)\\
		& Expert-Driven & 0.552 (0.023) & 0.534 (0.020)\\
		& Data-Driven & \textbf{0.607 (0.023)} & \textbf{0.567 (0.022)}\\
		\midrule
		\multirow{3}{*}{Short-Term} & None & 0.568 (0.017) & 0.572 (0.007)\\\
		& Expert-Driven & 0.586 (0.013) & 0.585 (0.009)\\
		& Data-Driven & \textbf{0.595 (0.013)} & \textbf{0.606 (0.009)}\\
		\bottomrule
	\end{tabular}
\end{table}
Our first set of experiments explores the utility of two-stage inference for data-derived motifs. Using the physiological data described in Section \ref{sec:cd}, we use the baseline data-derived motif discovery method, MDLats, to discover motifs using three different approaches to context: 
\begin{itemize}
	\item None: this represents the baseline motif discovery approach without context, described in Section \ref{sec:base} (\textit{i.e.}, MDLats). To faithfully reimplement MDLats, we discretized our data using the SAX representation technique \cite{lin2007experiencing}. 
	\item Expert-Driven: using two-stage inference to add context defined according to the domain rule discussed in Section \ref{sec:two_stage}. This rule is based on the clinically validated measure of glycemic variability, MAGE \cite{MAGE},
	\item Data-Driven: using two-stage inference to add context defined as the hidden state sequence in an HMM.
\end{itemize}
We use the learned motif representations to predict both hypo- and hyperglycemic events in the short and long term, using the pipeline discussed in \ref{sec:eval}. 

The results are presented in Table \ref{tab:dd_res}. The results of the baseline approach are poor (\textit{i.e.}, near random). Applied to the long-term task, the contextless baseline achieves an AUC=0.535 and AUC=0.527 for the tasks of predicting hypo- and hyperglycemic events respectively. But this level of predictive performance is not uncommon for such difficult tasks. Encouragingly, if we focus on the differences in performance, incorporating data-driven context leads to a significant increase in performance across \textit{both} tasks (AUC=0.607 and AUC=0.567). These differences are statistically significant at $\alpha=0.01$, as determined by a two-sided paired t test. For the short-term task performance is still poor. We hypothesize that this is because we do not discover many motifs that occur right before the prediction window. Upon inspection, only 18\% of our samples have a motif occurring in the 40 minutes prior to prediction. In a follow-up experiment we tested our data-driven contextual motif model on only those samples with a motif in the last 40 minutes, and achieved AUCs of 0.723 and 0.843 for hypo- and hyperglycemic prediction respectively. This demonstrates that for the short-term task, recent information is critical. 

\subsubsection{Data-Generating Motifs}
\label{sec:dgm}
\begin{table}
	\caption{Predictive utility of contextual data-generating motifs. We note two main trends in these results. First, the short-term task is much easier than the long-term task when using data-generating motifs. Second, CMMM consistently outperforms MMM.}
	\label{tab:dg_res}
	\begin{tabular}{ccll}
		\toprule
		& Method & \multicolumn{2}{c}{AUC (std)} \\
		\midrule
		Task & Model & Hypo & Hyper \\
		\midrule
		\multirow{2}{*}{Long-Term} &  MMM & 0.517 (0.023) & 0.588 (0.026)\\
		& CMMM & \textbf{0.539 (0.026)} & \textbf{0.595 (0.025)} \\
		\midrule
		\multirow{2}{*}{Short-Term} &  MMM & 0.811 (0.011) & 0.841 (0.007)\\
		& CMMM & \textbf{0.814 (0.010)} & \textbf{0.845 (0.007)}\\
		\bottomrule
	\end{tabular}
\end{table}
In our second set of experiments, we shift to data-generating motifs to test our joint inference approach. This allows us to measure the value of contextual motifs in a different setting (\textit{i.e.}, data-generating vs. data-derived). We compare MMM (presented in Section \ref{sec:base}) against our proposed CMMM on the glucose dataset. 

Discovery of data-generating motifs is more computationally intensive than discovering data-derived motifs. As such, we perform motif discovery on a randomly chosen subset of the data (40 days worth). After fitting our models, we apply maximum likelihood estimation to the remainder of the data to find motifs and context across all data, so as to leverage all training data. In a follow up experiment, we used an optimized GMM package \cite{pedregosa2011scikit} and the entire dataset. We did not, however, observe a significant improvement in predictive performance, suggesting that this subsampling has little effect on performance for this particular task. 

Based on a grid search over plausible values given our domain, we set the number of different motifs discovered to 20, the length of motifs $l_m=8$ (representing 40 minutes of glucose data), the length of each context window $l_c=72$, and the number of contexts to 2. We kept this setting constant across methods, and used the same evaluation metrics, namely AUC, as we did in Section \ref{sec:iiexp}. 

The results of our experiments are given in Table \ref{tab:dg_res}. Our proposed CMMM approach dominates the MMM approach, outperforming it on all tasks. The results directly compare the discriminative utility of the motifs discovered using MMM and CMMM

In Table \ref{tab:cmmm_context}, we examine the discriminative utility of the inferred context. Compared to motifs alone, motifs plus inferred context appears to do slightly worse. We hypothesize that a reason for the lack of performance improvement when adding context was that our pipeline failed to properly regularize when dimensionality was increased. Thus, we considered a third model in which we added noise to the motifs instead of context. Indeed, we see a decrease in performance when increasing dimensionality using {(motifs, noise)}, even though the actual information present remains constant. In all tasks except short-term hyperglycemic prediction, {(motifs, context)} outperforms {(motifs, noise)} ($\alpha=0.01$), suggesting there is some amount of discriminative information in the context.

\begin{table}
	\caption{Using motifs and context derived by the CMMM for predicting long- and short-term events. We observe best performance using {(motif)}, though {(motif, context)} consistently outperforms {(motif, noise)}.}
	\label{tab:cmmm_context}
	\begin{tabular}{ccll}
		\toprule
		& Method & \multicolumn{2}{c}{AUC (std)} \\
		\midrule
		Task & Representation & Hypo & Hyper \\
		\midrule
		\multirow{3}{*}{Long-Term} &  {(motif)} & 0.539 (0.026) & 0.595 (0.025) \\
		& {(motif, context)} & 0.533 (0.020) & 0.591 (0.029)\\
		&{(motif, noise)} &  0.510 (0.025) & 0.577 (0.023) \\
		\midrule
		\multirow{3}{*}{Short-Term} & {(motif)} & 0.814 (0.010) & 0.845 (0.007)\\
		& {(motif, context)} & 0.810 (0.009) & 0.843 (0.008)\\
		& {(motif, noise)} & 0.808 (0.010) & 0.843 (0.007)\\
		\bottomrule
	\end{tabular}
\end{table}
		
It is encouraging that the CMMM is able to discover higher quality motifs in addition to natively discovering contextual motifs. However, the failure of context to consistently improve performance in this task is troubling. We hypothesize this is due to having insufficient data (at the prediction stage, not motif discovery stage) to adequately compensate for the increased dimensionality that context introduces. Thus, to further test the efficacy of contextual motifs, we investigated performance on simulated data. 

\subsection{Experiments on Simulated Data}
To evaluate our proposed contextual motif methods across a range of settings, we turn to experiments using simulated data. Our simulated data are created using the CMMM generative model shown in Figure \ref{fig:cmmm}. We create labels for these data following the procedure given below, varying $\beta$ values to simulate different types of settings:
\begin{enumerate}
	\item For contextual motif $(\mathbf{m}, c) \in CM$, the set of all possible contextual motifs that can be generated under a CMMM:
	\begin{enumerate}
		\item draw a value for the contextual motif: $v_{(\mathbf{m},c)} \sim \mathbb{U}(-1, 1)$
	\end{enumerate}
	\item For each signal in the dataset $\mathbf{x} \in X$:
	\begin{enumerate}
		\item Compute a raw score for the signal: $s_\mathbf{x} = \sum_{(\mathbf{m},c) \in \mathbf{x}} v_{(\mathbf{m},c)}$
		\item Transform the score into a Bernoulli parameter: $$p_\mathbf{x} = \frac{\exp(\beta s_\mathbf{x})}{1+\exp(\beta s_\mathbf{x})}$$
		\item Draw the signal's outcome: $y_\mathbf{x} \sim \text{Ber}(p_\mathbf{x})$
	\end{enumerate}
\end{enumerate}

As an upper bound, we compare to two oracles that are based on the ground truth oracle:{(motifs)} and oracle:{(motifs, context)}. We compare {(motifs)} inferred by the CMMM, {(motifs, noise)}, two-stage:{(motifs, context)} (where we use the inferred motifs to derive the motif topic context as described in Section \ref{sec:two_stage}), and joint:{(motifs, context)}. We vary the importance of the contextual motifs by varying the value of $\beta$. This allows us to directly test the efficacy of our proposed methods. We examined predictive performance using 10,000 simulated signals, with identical CMMM hyperparameters to those used in our real data-experiments. We used the same evaluation pipeline as above, with the exception that the number of train-test splits is reduced to 25 since we had more data.

\begin{figure}
	\centering
	\includegraphics[width=\linewidth]{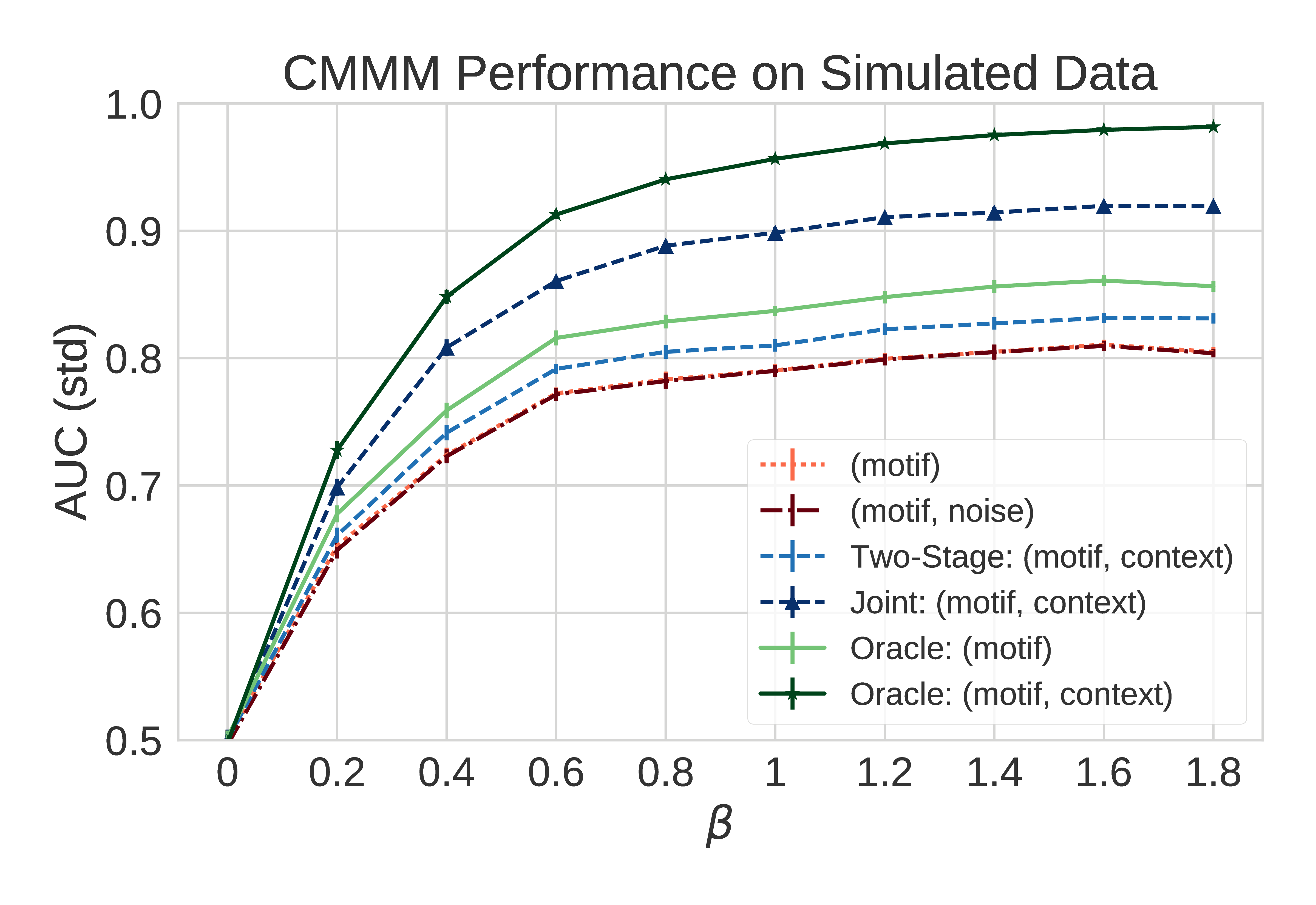}
	\caption{Here we see the efficacy of our proposed CMMM inference method on 10,000 simulated signals as we vary the strength of relationship between contextual motifs and outcome. Using only {(motifs)} results in suboptimal performance. Two-stage: {(motifs, context)}  improves performance, but not as much as using joint:{(motifs, context)}. Error bars on all lines signify 1 standard deviation.}
	\label{fig:cmmm_perf}
\end{figure}

The results are presented in Figure \ref{fig:cmmm_perf}. The {(motifs)} and {(motifs, noise)} representation perform the worst. The two-stage: {(motifs, context)} representation provides a minor boost to performance, and joint:{(motifs, context)} provides a major boost, surpassing the performance of the contextless oracle and approaching the performance of the oracle with context (significant with $\alpha=0.01$ when $\beta \ge 0.4$).

\section{Discussion}
\label{sec:disc}
Our results confirm both the importance of context in physiological time series and the ability of contextual motifs in capturing this context. Below, we summarize some of the main takeaways of our experiments and present a follow-up analysis in which we take a closer look at \text{what} the models actually learned. 

\textbf{Contextual Motifs vs. Motifs.} Our first set of experiments demonstrates that contextual motifs are more discriminative than their contextless counterparts, even when context is inferred. In Table \ref{tab:dd_res}, we showed that, particularly when using data-driven techniques, contextual motifs are more predictive of both hypo- and hyperglycemic events. These events are, on their face, opposites, as one involves an increase in glucose levels while the other involves a decrease. However, both events are indicative of poor glycemic control. The fact that our inferred context improves the predictive utility for motifs on both tasks suggests that contextual motifs are more informative of the physiological system than motifs alone.

\textbf{Data-Derived vs. Data-Generating.} Our second set of experiments, together with the first set, demonstrate the utility of considering both data-derived and data-generating motif frameworks. The data-derived motif discovery method, MDLats, was able to discover variable length motifs much faster than our data-generating motif discovery methods were able to discover fixed length motifs. However, we observe that the data-generating motif representation was much more useful for the short-term prediction task. In the hypoglycemic task, using the data-generating framework instead of the data-derived framework improved AUC from 0.607 to 0.814. The poor performance of the data-derived motif technique is due in large part to the lack of recent data. In a follow up experiment using the data-derived framework, we achieved an AUC of 0.723 for hypoglycemic prediction on a subset of the test set for which recent data were available. While detrimental for the short-term task, we saw that data-derived motifs can be helpful in other places. For instance, data-derived motifs outperformed data-generating in the long-term hypoglycemia task AUC 0.607 vs. 0.533. 

\textbf{Joint Inference vs. A Two-Stage Approach.} Both our second set of experiments in Section \ref{sec:dgm} and our experiments on simulated data suggest the utility of a joint inference approach. In our second experiment, the motifs discovered by the CMMM were more discriminative than those discovered by the MMM. Although we did not observe large differences between CMMM motifs with and without context in our real data, some of this appears to have resulted from regularization issues when increasing dimensionality due to insufficient training data. This is evidenced by the decline in performance when a noise context was added. In our large simulated dataset, we did not observe a performance difference between {(motifs)} and {(motifs, noise)}, suggesting that the regularization issues were a result of the data size and not our pipeline. Additionally, in our simulated experiment we observed a performance increase when we include context (from {(motifs)} to two-stage: {(motifs, context)}). Importantly, we observe a much larger performance increase when using the jointly inferred context (from two-stage: {(motifs, context)} to joint: {(motifs, context)}). These results suggest that joint inference allows for the discovery of both better motifs and better context. 

\begin{figure}
	\centering
	\includegraphics[width=0.8\linewidth]{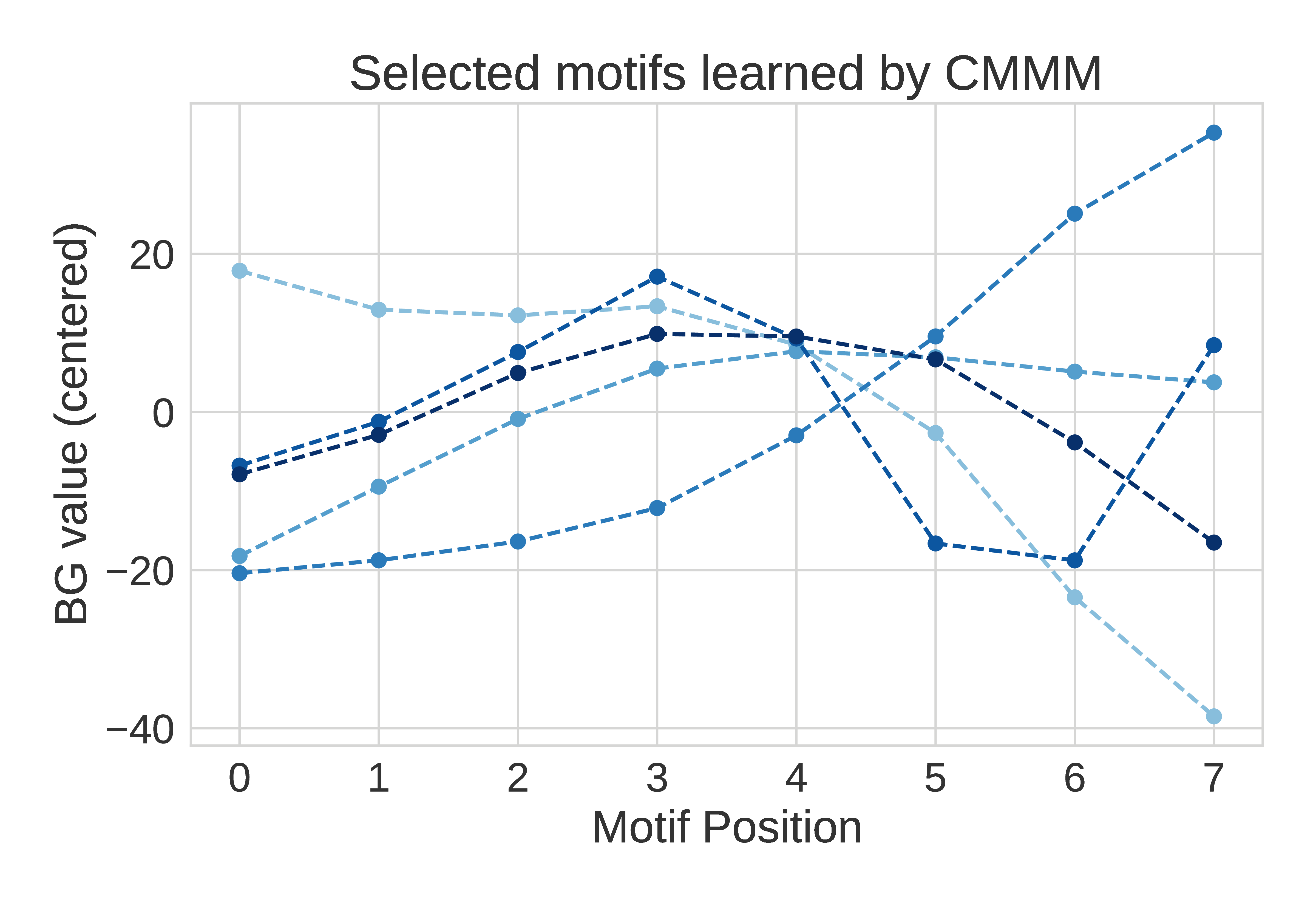}
	\caption{A hand-selected subset of motifs learned from the physiological data, selected to demonstrate the interesting range of motif morphologies present.}
	\label{fig:mots}
\end{figure}

\textbf{What are we learning?}
One of the main advantages of motifs is their interpretability. In Figure \ref{fig:mots}, we show the motifs we discovered in the CGM data. In this collection, we observe several interesting patterns, including nonlinear monotonic increases/decreases in glucose levels, demonstrating the nonlinearity of the glucoregulatory system. Additionally we observe `peak' and `trough' motifs, which appear indicative of hypo- and hyperglycemic events.

\begin{figure}
	\centering
	\includegraphics[width=0.8\linewidth]{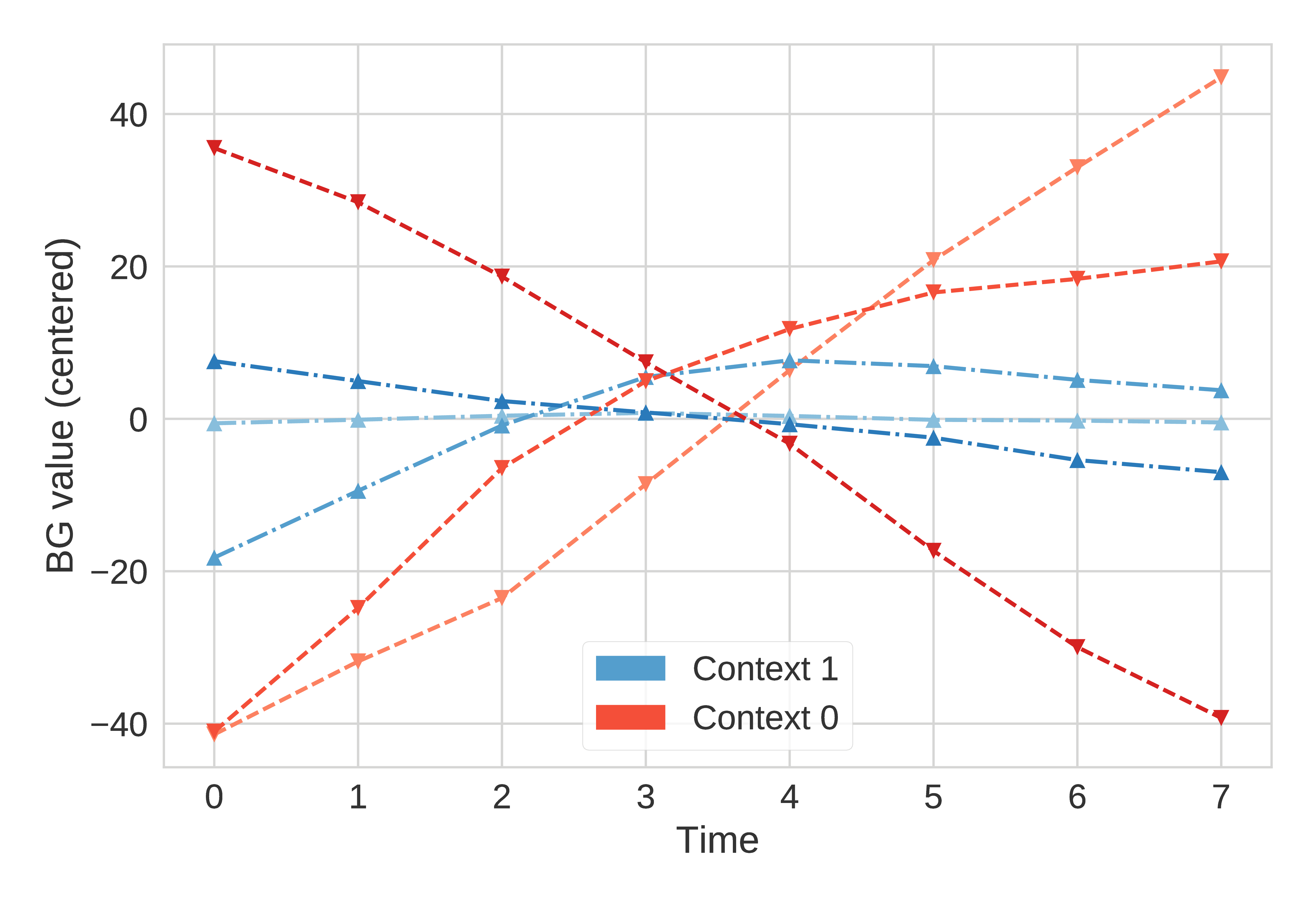}
	\caption{CMMM context comparison. We see that the top motifs under Context 0 have a much wider range of values than those under Context 1. This indicates that our learned context captures period of elevated glucose variability.}
	\label{fig:contexts}
\end{figure}

To examine the motifs and contexts we jointly inferred using the CMMM, we examine the difference in motif mixing probabilities under each context ($|\boldsymbol{\gamma}_0 - \boldsymbol{\gamma}_1|$). Here, the context we are learning represents inter-motif structure. The three motifs with the greatest negative and positive difference are shown in Figure \ref{fig:contexts}. These are the motifs which are the most indicative of the context under which they occur. Interestingly, we observe a clinically significant increase in the value range for motifs more likely under context 0 vs. 1. This suggests that context 0 represents periods of elevated glycemic variability, which could indicate the unsupervised, data-driven discovery of post/pre-meal contexts. 

\textbf{Limitations and Extensions.}
Our current contextual motif approach makes an important assumption: \textit{all contexts are categorical}. This makes sense for binary context sets, such as post/pre-meal. However, it may make sense to consider context along an ordinal scale. E.g., post-large meal, post-small meal, and fasting. One may consider extending the proposed framework to ordinal or even continuous representations. 

Our joint inference approach also assumes that the progression of context is memoryless, which may be unreasonable for certain physiological contexts (such as meals). This suggests a possible extension where context labels are generated by a stateful model.

Still, the utility of contextual motifs depends greatly on the context under consideration. Our joint inference method discovers a temporally varying motif topic context, which may or may not be useful depending on the task in question. When the context with the greatest potential utility is unobserved and difficult/impossible to derive from the signal, the utility of contextual motif discovery is limited. 

Contextual motifs do not currently take advantage of external labels to discover more discriminative motifs. By combining contextual motifs with existing techniques to improve motif performance, such as shapelets, it may be possible to further improve discriminative ability.

\section{Summary and Conclusions}
In this work, we described how the utility of motifs, as a representation for physiological time-series data, is limited when context is ignored. To address this limitation, we introduced the concept of \textit{contextual motifs} and proposed methods for contextual motif discovery.

We focused on the setting in which context is unobserved, since often one does not have access to contextual information. In this setting, we proposed a technique to jointly infer context and motifs.

We hypothesized that by incorporating context, we can discover more clinically meaningful motifs. To test this hypothesis, we ran a series of experiments on a CGM dataset of subjects with T1D. We considered two tasks: predicting hypo- and hyperglycemic events. Compared to a standard motif representation, our contextual motif discovery algorithms led to an increase in the AUC for both tasks. We conclude that contextual motifs are more discriminative than their contextless counterparts, even in the absence of observed context.

While contextual motifs can be discovered using a two-stage inference approach (Section \ref{sec:two_stage}), we hypothesized that one could learn higher quality motifs using a joint-inference approach. Our experiments on simulated data confirm this, though the difference is most pronounced when context plays an important role. We conclude that given a large dataset where context is relevant, CMMM is preferred over other approaches.

Contextual motifs present a new, richer, motif framework which augments motif representations to account for the context under which they occur. This new framework can be applied to sequential data in which context plays an important role, such as financial time series, energy-metering data, or physiological signals.

\section{Acknowledgments}
This work was supported by the National Science Foundation (NSF award numbers CNS-1330142 and IIS-1553146) and the National Institutes of Health (NIH research grant RO1 NIH/NHLBI-1R01HL102334-01). The views and conclusions in this document are those of the authors and should not be interpreted as necessarily representing the official policies, either expressed or implied, of the NSF or the NIH.
\appendix

\bibliographystyle{abbrv}
\bibliography{references}

%% file: contextual_motifs.bbl
\begin{thebibliography}{10}

\bibitem{MAGE}
P.~A. Baghurst.
\newblock Calculating the mean amplitude of glycemic excursion from continuous
  glucose monitoring data: an automated algorithm.
\newblock {\em Diabetes technology \& therapeutics}, 13(3):296--302, 2011.

\bibitem{bailey_fitting_1994}
T.~L. Bailey and C.~Elkan.
\newblock Fitting a mixture model by expectation maximization to discover
  motifs in biopolymers.
\newblock {\em Proceedings. International Conference on Intelligent Systems for
  Molecular Biology}, 2:28--36, 1994.

\bibitem{bergenstal2015glycemic}
R.~M. Bergenstal.
\newblock Glycemic variability and diabetes complications: Does it matter?
  simply put, there are better glycemic markers!
\newblock {\em Diabetes care}, 38(8):1615--1621, 2015.

\bibitem{blei2003latent}
D.~M. Blei, A.~Y. Ng, and M.~I. Jordan.
\newblock Latent {Dirichlet} allocation.
\newblock {\em the Journal of machine Learning research}, 3:993--1022, 2003.

\bibitem{breton2008analysis}
M.~Breton and B.~Kovatchev.
\newblock Analysis, modeling, and simulation of the accuracy of continuous
  glucose sensors.
\newblock {\em Journal of Diabetes Science and Technology}, 2(5):853--862,
  2008.

\bibitem{buhler2002finding}
J.~Buhler and M.~Tompa.
\newblock Finding motifs using random projections.
\newblock {\em Journal of computational biology}, 9(2):225--242, 2002.

\bibitem{cheng_discriminative_2007}
H.~Cheng, X.~Yan, J.~Han, and C.~W. Hsu.
\newblock Discriminative {Frequent} {Pattern} {Analysis} for {Effective}
  {Classification}.
\newblock In {\em 2007 {IEEE} 23rd {International} {Conference} on {Data}
  {Engineering}}, pages 716--725, Apr. 2007.

\bibitem{chiu2003probabilistic}
B.~Chiu, E.~Keogh, and S.~Lonardi.
\newblock Probabilistic discovery of time series motifs.
\newblock In {\em Proceedings of the ninth ACM SIGKDD international conference
  on Knowledge discovery and data mining}, pages 493--498. ACM, 2003.

\bibitem{chui2014introduction}
C.~K. Chui.
\newblock {\em An introduction to wavelets}, volume~1.
\newblock Academic press, 2014.

\bibitem{grabocka2016latent}
J.~Grabocka, N.~Schilling, and L.~Schmidt-Thieme.
\newblock Latent time-series motifs.
\newblock {\em ACM Transactions on Knowledge Discovery from Data (TKDD)},
  11(1):6, 2016.

\bibitem{hoffman2014no}
M.~D. Hoffman and A.~Gelman.
\newblock The no-u-turn sampler: adaptively setting path lengths in hamiltonian
  monte carlo.
\newblock {\em Journal of Machine Learning Research}, 15(1):1593--1623, 2014.

\bibitem{jaiswal2014association}
M.~Jaiswal, K.~McKeon, N.~Comment, J.~Henderson, S.~Swanson, C.~Plunkett,
  P.~Nelson, and R.~Pop-Busui.
\newblock Association between impaired cardiovascular autonomic function and
  hypoglycemia in patients with type 1 diabetes.
\newblock {\em Diabetes care}, 37(9):2616--2621, 2014.

\bibitem{jensen2006generic}
K.~L. Jensen, M.~P. Styczynski, I.~Rigoutsos, and G.~N. Stephanopoulos.
\newblock A generic motif discovery algorithm for sequential data.
\newblock {\em Bioinformatics}, 22(1):21--28, 2006.

\bibitem{kleiger1987decreased}
R.~E. Kleiger, J.~P. Miller, J.~T. Bigger, and A.~J. Moss.
\newblock Decreased heart rate variability and its association with increased
  mortality after acute myocardial infarction.
\newblock {\em The American journal of cardiology}, 59(4):256--262, 1987.

\bibitem{lin2002finding}
J.~Lin, E.~Keogh, S.~Lonardi, and P.~Patel.
\newblock Finding motifs in time series.
\newblock In {\em Proc. of the 2nd Workshop on Temporal Data Mining}, pages
  53--68, 2002.

\bibitem{lin2007experiencing}
J.~Lin, E.~Keogh, L.~Wei, and S.~Lonardi.
\newblock Experiencing sax: a novel symbolic representation of time series.
\newblock {\em Data Mining and knowledge discovery}, 15(2):107--144, 2007.

\bibitem{lin2008baycis}
T.-h. Lin, P.~Ray, G.~K. Sandve, S.~Uguroglu, and E.~P. Xing.
\newblock Baycis: a {Bayesian} hierarchical hmm for cis-regulatory module
  decoding in metazoan genomes.
\newblock In {\em Annual International Conference on Research in Computational
  Molecular Biology}, pages 66--81. Springer, 2008.

\bibitem{liu2015efficient}
B.~Liu, J.~Li, C.~Chen, W.~Tan, Q.~Chen, and M.~Zhou.
\newblock Efficient motif discovery for large-scale time series in healthcare.
\newblock {\em IEEE Transactions on Industrial Informatics}, 11(3):583--590,
  2015.

\bibitem{liu1996metropolized}
J.~S. Liu.
\newblock Metropolized gibbs sampler: an improvement.
\newblock Technical report, Technical report, Dept. Statistics, Stanford Univ,
  1996.

\bibitem{minnen2006discovering}
D.~Minnen, T.~Starner, I.~Essa, and C.~Isbell.
\newblock Discovering characteristic actions from on-body sensor data.
\newblock In {\em 2006 10th IEEE international symposium on wearable
  computers}, pages 11--18. IEEE, 2006.

\bibitem{molnar1970mean}
G.~D. Molnar, J.~W. Rosevear, E.~Ackerman, L.~C. Gatewood, W.~F. Taylor, et~al.
\newblock Mean amplitude of glycemic excursions, a measure of diabetic
  instability.
\newblock {\em Diabetes}, 19(9):644--655, 1970.

\bibitem{mueen2014time}
A.~Mueen.
\newblock Time series motif discovery: dimensions and applications.
\newblock {\em Wiley Interdisciplinary Reviews: Data Mining and Knowledge
  Discovery}, 4(2):152--159, 2014.

\bibitem{muggeo2000fasting}
M.~Muggeo, G.~Zoppini, E.~Bonora, E.~Brun, R.~C. Bonadonna, P.~Moghetti, and
  G.~Verlato.
\newblock Fasting plasma glucose variability predicts 10-year survival of type
  2 diabetic patients: the verona diabetes study.
\newblock {\em Diabetes care}, 23(1):45--50, 2000.

\bibitem{murphy2002dynamic}
K.~P. Murphy.
\newblock Dynamic bayesian networks.
\newblock {\em Probabilistic Graphical Models, M. Jordan}, 7, 2002.

\bibitem{nakamura2016multiscale}
T.~Nakamura, K.~Kiyono, H.~Wendt, P.~Abry, and Y.~Yamamoto.
\newblock Multiscale analysis of intensive longitudinal biomedical signals and
  its clinical applications.
\newblock {\em Proceedings of the IEEE}, 104(2):242--261, 2016.

\bibitem{needleman1970general}
S.~B. Needleman and C.~D. Wunsch.
\newblock A general method applicable to the search for similarities in the
  amino acid sequence of two proteins.
\newblock {\em Journal of molecular biology}, 48(3):443--453, 1970.

\bibitem{oates2002peruse}
T.~Oates.
\newblock Peruse: An unsupervised algorithm for finding recurring patterns in
  time series.
\newblock In {\em Data Mining, 2002. ICDM 2003. Proceedings. 2002 IEEE
  International Conference on}, pages 330--337. IEEE, 2002.

\bibitem{pedregosa2011scikit}
F.~Pedregosa, G.~Varoquaux, A.~Gramfort, V.~Michel, B.~Thirion, O.~Grisel,
  M.~Blondel, P.~Prettenhofer, R.~Weiss, V.~Dubourg, et~al.
\newblock Scikit-learn: Machine learning in python.
\newblock {\em Journal of Machine Learning Research}, 12(Oct):2825--2830, 2011.

\bibitem{ranu_graphsig:_2009}
S.~Ranu and A.~K. Singh.
\newblock {GraphSig}: {A} {Scalable} {Approach} to {Mining} {Significant}
  {Subgraphs} in {Large} {Graph} {Databases}.
\newblock In {\em 2009 {IEEE} 25th {International} {Conference} on {Data}
  {Engineering}}, pages 844--855, Mar. 2009.

\bibitem{salvatier2016probabilistic}
J.~Salvatier, T.~V. Wiecki, and C.~Fonnesbeck.
\newblock Probabilistic programming in python using pymc3.
\newblock {\em PeerJ Computer Science}, 2:e55, 2016.

\bibitem{saria_discovering_2011}
S.~Saria, A.~Duchi, and D.~Koller.
\newblock Discovering deformable motifs in continuous time series data.
\newblock In {\em {IJCAI} {Proceedings}-{International} {Joint} {Conference} on
  {Artificial} {Intelligence}}, volume~22, page 1465, 2011.

\bibitem{saria2010learning}
S.~Saria, D.~Koller, and A.~Penn.
\newblock Learning individual and population level traits from clinical
  temporal data.
\newblock In {\em Proc. Neural Information Processing Systems (NIPS),
  Predictive Models in Personalized Medicine workshop}. Citeseer, 2010.

\bibitem{shokoohi2015discovery}
M.~Shokoohi-Yekta, Y.~Chen, B.~Campana, B.~Hu, J.~Zakaria, and E.~Keogh.
\newblock Discovery of meaningful rules in time series.
\newblock In {\em Proceedings of the 21th ACM SIGKDD International Conference
  on Knowledge Discovery and Data Mining}, pages 1085--1094. ACM, 2015.

\bibitem{siddharthan2005phylogibbs}
R.~Siddharthan, E.~D. Siggia, and E.~Van~Nimwegen.
\newblock Phylogibbs: a {Gibbs} sampling motif finder that incorporates
  phylogeny.
\newblock {\em PLoS Comput Biol}, 1(7):e67, 2005.

\bibitem{syed2010motif}
Z.~Syed, C.~Stultz, M.~Kellis, P.~Indyk, and J.~Guttag.
\newblock Motif discovery in physiological datasets: a methodology for
  inferring predictive elements.
\newblock {\em ACM Transactions on Knowledge Discovery from Data (TKDD)},
  4(1):2, 2010.

\bibitem{vahdatpour2009toward}
A.~Vahdatpour, N.~Amini, and M.~Sarrafzadeh.
\newblock Toward unsupervised activity discovery using multi-dimensional motif
  detection in time series.
\newblock In {\em IJCAI}, volume~9, pages 1261--1266, 2009.

\bibitem{van2015learning}
A.~Van~Esbroeck.
\newblock {\em Learning Better Clinical Risk Models}.
\newblock PhD thesis, The University of Michigan, 2015.

\bibitem{van2012heart}
A.~Van~Esbroeck, C.-C. Chia, and Z.~Syed.
\newblock Heart rate topic models.
\newblock In {\em The Twenty-Sixth AAAI Conference on Artificial Intelligence},
  volume 1001, page 48109, 2012.

\bibitem{weinzimer_fully_2008}
S.~A. Weinzimer, G.~M. Steil, K.~L. Swan, J.~Dziura, N.~Kurtz, and W.~V.
  Tamborlane.
\newblock Fully automated closed-loop insulin delivery versus semiautomated
  hybrid control in pediatric patients with type 1 diabetes using an artificial
  pancreas.
\newblock {\em Diabetes care}, 31(5):934--939, 2008.

\bibitem{ye2009time}
L.~Ye and E.~Keogh.
\newblock Time series shapelets: a new primitive for data mining.
\newblock In {\em Proceedings of the 15th ACM SIGKDD international conference
  on Knowledge discovery and data mining}, pages 947--956. ACM, 2009.

\bibitem{zhou2004cismodule}
Q.~Zhou and W.~H. Wong.
\newblock Cismodule: de novo discovery of cis-regulatory modules by
  hierarchical mixture modeling.
\newblock {\em Proceedings of the national academy of sciences of the United
  States of America}, 101(33):12114--12119, 2004.

\bibitem{zoungas2010severe}
S.~Zoungas, A.~Patel, J.~Chalmers, B.~E. de~Galan, Q.~Li, L.~Billot,
  M.~Woodward, T.~Ninomiya, B.~Neal, S.~MacMahon, et~al.
\newblock Severe hypoglycemia and risks of vascular events and death.
\newblock {\em New England Journal of Medicine}, 363(15):1410--1418, 2010.

\end{thebibliography}
